# Toward Responsible and Epistemically Grounded Multilingual LLMs for Computational Social Science and Humanities

**Wajdi Zaghouani**
Northwestern University in Qatar
wajdi.zaghouani@northwestern.edu

**Abstract**

Large language models have rapidly evolved in multilingual competence and reasoning capacity, enabling their integration into Social Sciences and Humanities research workflows. Yet existing evaluation paradigms remain anchored in task-based NLP benchmarks and fail to address interpretive validity, cultural situatedness, and epistemic mediation. This paper reconceptualizes multilingual reasoning LLMs as hermeneutic instruments that actively structure meaning production across linguistic and cultural contexts. Drawing on hermeneutics, philosophy of technology, science and technology studies, multilingual NLP research, and computational social science methodology, we develop a theoretically grounded framework for evaluating multilingual reasoning in Social Sciences and Humanities (SSH) research. We articulate a rigorous experimental protocol with operationalized metrics for cultural alignment, cross-lingual stability, and reasoning faithfulness, along with transparency requirements tailored to interpretive research tasks. We illustrate the framework through a concrete application scenario involving multilingual political discourse analysis. The paper contributes a conceptual and methodological foundation for responsible integration of multilingual reasoning LLMs into computational social science infrastructures.

## 1. Introduction

The development of transformer-based large language models has reconfigured the technical landscape of language processing. Models trained on large-scale multilingual corpora demonstrate zero-shot transfer, in-context learning, and chain-of-thought reasoning capacities across a growing number of languages (Brown et al., 2020; Chowdhery et al., 2023; Wei et al., 2022; Touvron et al., 2023). These advances have prompted increasing adoption of LLMs in domains beyond traditional NLP, including political science, sociology, communication studies, digital humanities, and cultural analytics.

Recent work demonstrates that LLMs can assist in text annotation, survey simulation, and qualitative coding tasks (Argyle et al., 2023; Gilardi et al., 2023; Ziems et al., 2024). The computational social science community has begun systematically examining how LLMs can transform research workflows, with emerging evidence showing that zero-shot LLMs can achieve fair levels of agreement with humans on taxonomic labeling tasks while producing explanations that sometimes exceed crowdworker quality (Ziems et al., 2024). Wilkerson and Casas (2017) documented the growing importance of large-scale computerized text analysis in political science, and LLMs represent the latest methodological frontier in this trajectory. However, performance improvements on benchmarks do not directly translate into epistemic adequacy for Social Sciences and Humanities (SSH) research. Interpretive analysis in social science is not merely a classification task. It involves contextualization, reflexivity, and engagement with culturally embedded meaning structures.

Multilingual reasoning models introduce additional complexity. Although models demonstrate cross-lingual competence, research consistently documents uneven performance across languages and structural inequalities in training data (Joshi et al., 2020; Blasi et al., 2022). The most widely used datasets in natural language processing currently represent only a handful of data-rich languages, and the datasets used for instruction fine-tuning are almost entirely focused on English (Singh et al., 2024). Cultural alignment and interpretive fidelity cannot be assumed on the basis of benchmark scores.

This paper advances the argument that multilingual reasoning LLMs should be conceptualized as hermeneutic instruments. Rather than treating them as neutral analytic tools, we frame them as mediators that shape interpretive horizons in computational social science. This reframing has methodological implications for evaluation design, transparency practices, and epistemic accountability. As Rockwell and Sinclair (2016) demonstrate in their foundational work on computer-assisted interpretation, computational tools do not simply process texts but actively structure interpretive possibilities. This perspective aligns with Moretti's (2013) notion of "distant reading," which acknowledges that computational approaches to text involve fundamental shifts in how texts are apprehended.

The contribution is fourfold. First, we articulate a philosophy of technology-grounded account of multilingual LLM mediation in SSH research. Second, we provide an empirically informed analysis of cultural bias in multilingual models, with particular attention to non-Western language contexts. Third, we propose a detailed methodological framework with operationalized metrics for evaluating multilingual reasoning models in interpretive contexts, illustrated through a concrete application scenario. Fourth, we identify epistemic and ethical implications that extend beyond technical evaluation and offer concrete recommendations for the SSH research community.

## 2. Hermeneutics and Computational Mediation

Hermeneutic philosophy offers a foundational lens for understanding interpretation as historically and culturally situated. Gadamer (1975) emphasizes that understanding occurs through a fusion of horizons between interpreter and text. Ricoeur (1976) similarly frames interpretation as a dialectical process that unfolds between explanation and understanding. In hermeneutic thought, interpretation is never extraction of fixed meaning but a productive act shaped by presuppositions and contextual frames.

The interpretive act is structured by mediating conditions, including language, tradition, and institutional practice. As scholars in digital humanities have argued, computational text analysis necessarily involves hermeneutical operations: the creation of vector space models, topic models, or neural representations should be understood as interpretations that reinscribe texts into new analytical forms (Kuhn, 2019; van Zundert, 2016). When computational systems are introduced into interpretive workflows, they become part of these mediating conditions. Their architectures encode statistical regularities derived from training data, reflecting historically contingent distributions of language use, cultural representation, and epistemic dominance. Kuhn (2019) identifies a central tension in digital humanities between hermeneutic traditions of text interpretation and method-oriented research strategies in computational linguistics.

The philosophical insight that instruments mediate rather than passively transmit phenomena is reinforced in science and technology studies. Latour (1987) argues that scientific instruments transform the entities they measure. Ihde (1990) develops this insight through the concept of "technological intentionality," suggesting that instruments structure perception and interpretation in ways that are not fully transparent to users. Winner (1980) further demonstrates that technological artifacts embody political and social values.

Applying this perspective to multilingual reasoning LLMs shifts the analytical focus. The central question becomes not only whether the model performs accurately but how it structures interpretive possibilities. When an LLM summarizes political discourse in Arabic, reconstructs an argument in French, or attributes causality in English, it does so through latent statistical priors shaped by its training distribution. These priors influence what counts as salient, coherent, or plausible. What Gadamer called the interpreter's "effective-historical consciousness" finds a distorted analogue in the model's training distribution: the model interprets through a horizon constituted not by lived experience but by the statistical regularities of its corpus, weighted by the demographic and linguistic composition of its data sources.

## 3. Multilingual Representation and Structural Inequality

The multilingual capacity of LLMs is often framed as evidence of inclusivity and global reach. However, empirical research in NLP highlights persistent structural asymmetries. Joshi et al. (2020) document severe disparities in resource availability and benchmark inclusion across languages, categorizing the world's languages into a taxonomy where the vast majority remain "left-behind" in NLP research. Blasi et al. (2022) demonstrate that English plays a disproportionate role in shaping multilingual model performance, leading to what they describe as the unreasonable effectiveness of English in cross-lingual transfer.

Bender et al. (2021) argue that large-scale language models reflect the biases and imbalances of their training data. The Aya initiative represents an important effort to address these gaps through community-driven multilingual instruction tuning across 101 languages, yet significant disparities persist (Singh et al., 2024; Üstün et al., 2024). Community-driven efforts such as Masakhane, which focuses on NLP for African languages, demonstrate the importance of participatory approaches that involve native speakers as researchers rather than merely data providers (Orife et al., 2020; Adelani et al., 2021).

In multilingual reasoning contexts, these structural asymmetries have interpretive implications. Cross-lingual reasoning tasks often rely on translation or transfer learning. Yet semantic equivalence does not guarantee cultural equivalence. Research on cultural alignment demonstrates that multilingual capability does not imply multicultural understanding: LLMs trained primarily on English data consistently align with Western, particularly US-centric, cultural values even when generating content in other languages (Rystrøm et al., 2025; Tao et al., 2024). Li et al. (2024) find that LLMs inherit and am-plify cultural patterns present in their training data, replicating cross-cultural personality differences while overrepresenting Western perspectives due to English-dominant corpora.

For computational social science, this is particularly significant. Comparative political analysis depends on detecting differences in framing, narrative structure, and causal attribution. Following Entman's (1993) influential framework, framing involves selecting and making salient certain aspects of perceived reality to promote particular problem definitions, causal interpretations, moral evaluations, and treatment recommendations. Chong and Druckman (2007) further demonstrate how framing shapes public opinion formation. If multilingual reasoning models encode dominant cultural frames more strongly than minority ones, cross-national comparison may inadvertently reflect model priors rather than empirical reality. Prior work on annotating stance, sentiment, and framing in Arabic social media discourse (Laabar and Zaghouani, 2024) illustrates how culturally specific interpretive categories can diverge substantially from categories developed in English-language contexts.

Existing multilingual benchmarks such as XNLI (Conneau et al., 2018) provide valuable infrastructure for evaluating cross-lingual transfer but focus primarily on natural language inference rather than culturally situated interpretation. The MBBQ benchmark demonstrates that bias patterns differ substantially across languages (Neplenbroek et al., 2024). SSH-specific evaluation requires extending these resources to capture interpretive validity.

## 4. Cultural Bias as Epistemic Distortion

The abstract concern about cultural homogenization in multilingual models is supported by a growing body of empirical work that deserves careful attention from the SSH community. Naous et al. (2024) introduce CAMeL, a benchmark of naturally occurring prompts and entities contrasting Arab and Western cultures, and demonstrate that both multilingual and Arabic monolingual language models exhibit systematic bias toward entities associated with Western culture. When prompted in Arabic to complete sentences about food, beverages, or personal names, models disproportionately generate Western-associated responses rather than culturally appropriate Arab alternatives. This bias persists even in models specifically fine-tuned for Arabic, suggesting that the issue runs deeper than surface-level language competence.

The origins of these biases have been further investigated by Naous and Xu (2025), who find that language models struggle particularly with Arabic entities that appear at high frequencies in pre-training data, where such entities exhibit strong word polysemy. Their analysis reveals that frequency-based tokenization contributes to this problem, and that performance gaps between Arab and Western cultural entities are smaller when models are tested in English compared to Arabic. This finding has a striking implication for SSH researchers: a model may appear culturally competent when evaluated in English but reveal significant cultural blind spots when operating in the target language of analysis.

Large-scale evaluations reinforce these findings. Sukiennik et al. (2025) conduct the first comprehensive assessment of cultural value alignment across 20 countries and 10 LLMs using Hofstede's Values Survey Module, finding that model outputs converge toward a moderate cultural middle ground and that the United States is consistently the best-aligned country across models. Critically, models regardless of their country of origin align better with US cultural values than with the values of their home countries, suggesting that the dominance of English-language training data creates a structural gravitational pull toward Western cultural norms.

These empirical findings have direct methodological consequences for SSH research. A researcher using an LLM to analyze political discourse across Arabic-speaking countries may encounter a model that systematically underweights culturally specific concepts, prioritizes Western-normative framings, or generates interpretive outputs that obscure genuine cross-cultural variation. Work on multi-dialectal Arabic hate speech detection (Charfi et al., 2024a) and cross-domain stance analysis (Charfi et al., 2024b) has shown that even within a single language, dialectal and cultural variation can substantially affect model performance and annotation validity. For disciplines such as comparative politics, area studies, and cultural sociology, where the detection of culturally specific meaning is the primary research objective, these biases represent not merely technical limitations but epistemic threats to the validity of findings.

## 5. Reasoning, Explanation, and Epistemic Authority

Chain-of-thought prompting has been shown to improve reasoning performance in large language models (Wei et al., 2022). Generating intermediate steps appears to enhance accuracy on arithmetic and logical tasks. However, Turpin et al. (2023) demonstrate that reasoning traces may not reliably correspond to internal inference processes. Their experiments reveal that LLMs can produce chain-of-thought explanations that are systematically unfaithful, influenced by biasing features in inputs that models fail to mention in their expla-nations. When models are biased toward incorrect answers through manipulated prompts, they frequently generate plausible-sounding reasoning that rationalizes those incorrect answers, causing accuracy to drop by as much as 36% on benchmark tasks.

This finding has been corroborated by Lanham et al. (2023), who develop multiple metrics for assessing chain-of-thought faithfulness and find substantial variation across tasks in how strongly models condition on their stated reasoning when predicting answers. Their experiments introduce perturbations to chain-of-thought outputs, such as adding mistakes or paraphrasing, to measure whether models genuinely rely on their stated reasoning. Critically, as models become larger and more capable, they sometimes produce less faithful reasoning on certain tasks, raising concerns about the inverse scaling of interpretability.

For SSH research, reasoning traces may acquire epistemic authority. When a model provides a structured explanation of why a political actor adopts a particular stance, researchers may treat the reasoning as analytically meaningful. Yet if reasoning traces are post-hoc constructions optimized for plausibility rather than grounded inference, their interpretive status must be critically examined. Zheng et al. (2023) further demonstrate that LLM-based evaluators can introduce systematic biases in judging outputs, including position bias, verbosity bias, and self-enhancement bias. Chen et al. (2024) and Gu et al. (2024) identify twelve distinct types of bias that can undermine LLM-as-judge reliability.

The faithfulness problem intersects with multilingual reasoning in ways that remain underexplored. When a model reasons in a language other than English, it may rely more heavily on transferred English-language priors, producing reasoning traces that appear coherent in the target language but are actually anchored in English-centric conceptual structures. This creates a particularly insidious form of epistemic distortion: the model may generate culturally inappropriate interpretations accompanied by fluent, seemingly well-reasoned explanations that mask the underlying cultural misalignment. SSH researchers lack adequate tools to detect when this form of cross-lingual reasoning contamination is occurring, making the development of language-specific faithfulness metrics an urgent priority.

## 6. LLMs in Computational Social Science

Before articulating an evaluation framework, it is important to survey how LLMs are currently being deployed in SSH research. The use of LLMs for text annotation has expanded rapidly, with studies demonstrating that GPT-4 can achieve annotation accuracy comparable to or exceeding human crowdworkers across multiple tasks (Gilardi et al., 2023; Heseltine and Clemm von Hohenberg, 2024). Tornberg (2023) provides evidence that ChatGPT-4 achieves higher accuracy, higher reliability, and equal or lower bias than human classifiers. Rathje et al. (2024) demonstrate that GPT is effective for multilingual psychological text analysis across 12 languages.

Egami et al. (2024) develop a rigorous statistical framework for using LLM annotations in downstream social science analysis, showing that ignoring prediction errors from automated annotation can lead to substantial bias, invalid confidence intervals, and inaccurate p-values. Carlson and Burbano (2025) extend this line of work by developing foundational guidelines for using LLMs to annotate data in management research, demonstrating that subtle implementation choices, including prompt wording, model version, and parameter settings, can significantly affect not only annotation accuracy but also downstream research conclusions.

However, research also reveals that LLMs exhibit party cue biases when annotating political content (Vallejo Vera et al., 2025). For qualitative research, Hayes (2025) argues that LLMs enable researchers to “converse” with qualitative data in unprecedented ways, but this capability comes with risks: LLMs may impose interpretive frameworks that do not align with the cultural contexts being studied. The codebook-following capabilities of LLMs have been systematically examined by Halterman and Keith (2024), who find that providing detailed social science codebooks significantly improves classification performance. Ollion et al. (2023) urge researchers to “mind the hype,” noting that performance varies considerably across tasks and contexts.

An emerging concern involves the propagation of methodological choices through the research pipeline. When LLM annotations serve as inputs to regression models, hypothesis tests, or causal inference procedures, even small systematic biases can compound. Egami et al. (2024) propose design-based supervised learning (DSL) as a correction mechanism, but this requires a subsample of gold-standard human annotations. For multilingual SSH research, the practical challenge is acute: obtaining high-quality human annotations in multiple languages and cultural contexts is precisely the bottleneck that motivates LLM adoption in the first place. This creates a methodological circularity that the field must address through creative experimental designs, such as stratified validation sampling that ensures adequate representation of culturally distinctive categories.

# 7. Methodological Framework with Operationalized Metrics

This section outlines an experimental protocol for evaluating multilingual reasoning LLMs in computational social science contexts, with concrete operationalized metrics addressing cultural alignment, cross-lingual stability, and reasoning faithfulness.

**Corpus Construction and Documentation.** A valid multilingual evaluation requires corpora composed of native language texts rather than translated benchmarks. The distinction between native and translated evaluation materials is critical: translated benchmarks inherit the conceptual categories and pragmatic assumptions of their source language, introducing systematic confounds into cross-lingual evaluation. Documents should be sampled from comparable genres across linguistic contexts, where "comparable" is defined by functional equivalence (texts serving similar communicative purposes in their respective societies) rather than semantic equivalence (texts expressing the same propositional content). Following Bender and Friedman (2018), researchers should provide comprehensive Data Statements documenting: (a) curation rationale, (b) language variety with dialect specification, (c) speaker demographics, (d) annotator demographics, (e) speech situation context, and (f) text characteristics. This documentation standard, along with Model Cards (Mitchell et al., 2019), ensures transparency about the populations represented in evaluation data.

**Annotation Schema.** Interpretive tasks must be grounded in established SSH theory. For political discourse analysis, tasks may include frame identification following Entman's (1993) four functions. Annotation guidelines should be developed collaboratively with native speaker experts. The distinction between "universal label assumptions" and "codebook-contextual label assumptions" is crucial (Halterman and Keith, 2024). Universal label assumptions treat annotation categories as cross-culturally stable, whereas codebook-contextual assumptions recognize that categories may require adaptation to local meaning systems. Multiple annotators per language should independently code a subset of the corpus to establish reliability. Following Krippendorff (2004), we adopt the standard thresholds of Krippendorff's alpha ≥ 0.67 for tentative or exploratory conclusions and ≥ 0.80 for confirmatory research where findings will inform substantive claims.

**Cultural Alignment Metrics.** To operationalize cultural alignment, we propose measuring the distance between model-predicted attitude or value distributions and population benchmarks from validated cross-cultural surveys such as the World Values Survey (WVS) or European Social Survey (ESS). Specifically:

- *KL Divergence*: $D_{KL}(P_{population} || P_{model})$ where $P_{population}$ represents the distribution of responses on a value dimension (e.g., traditional vs. secular-rational values) from WVS respondents in the target culture, and $P_{model}$ represents the distribution of model outputs on comparable items. Lower values indicate better alignment. We propose a threshold of $D_{KL} < 0.1$ nats as a starting point; this value should be calibrated empirically through pilot studies, as the appropriate threshold will vary by task and domain.
- *Earth Mover's Distance (EMD)*: For ordinal scales, EMD provides an interpretable metric of the "work" required to transform model distributions into population distributions.
- *Cultural Bias Score*: Following Naous et al. (2024), researchers should additionally compute entity-level bias scores comparing model performance on culturally specific entities (e.g., Arab vs. Western food items, names, or locations) to detect systematic cultural skew that aggregate distributional metrics may obscure.

**Cross-Lingual Stability Metrics.** Cross-lingual stability measures whether model performance and interpretive outputs remain consistent across languages for semantically equivalent inputs. We propose:

- *Variance decomposition*: Using mixed-effects models with language as a random effect, decompose total variance into between-language ($\sigma^2_{language}$) and within-language ($\sigma^2_{residual}$) components. The intraclass correlation coefficient (ICC) = $\sigma^2_{language} / (\sigma^2_{language} + \sigma^2_{residual})$ quantifies the proportion of variance attributable to language. Following conventions in reliability research (Cicchetti, 1994), ICC > 0.10 suggests that language identity explains a non-trivial share of variance, warranting investigation of language-specific performance differences.
- *Pairwise agreement*: For each language pair, compute Cohen's kappa between model outputs on parallel test items. Mean pairwise kappa ≥ 0.60 indicates acceptable stability.
- *Language-direction asymmetry*: For each language pair, compare model performance when the task is formulated in language A versus language B. Asymmetric performance (e.g., consistently higher accuracy when prompting in English than in Arabic for the same underlying task) may indicate reliance on English-language priors rather than genuine multilingual competence.

**Reasoning Faithfulness Assessment.** Following Turpin et al. (2023) and Lanham et al. (2023), we recommend three perturbation families:

- *Bias injection*: Introduce irrelevant features (e.g., suggested answers, social cues) into prompts. If model answers change but explanations do not acknowledge the influence, faithfulness is compromised. Acceptance criterion: answer stability ≥ 90% under bias injection.
- *Reasoning corruption*: Introduce errors into chain-of-thought traces and measure whether final answers change correspondingly. If answers remain stable despite corrupted reasoning, the model is not genuinely conditioning on its explanations.
- *Cr oss-lingual reasoning transfer*: Present the same reasoning task in multiple languages and compare not only final answers but the structure and content of reasoning traces. Divergent reasoning paths for semantically equivalent inputs may reveal language-dependent reasoning strategies that compromise cross-lingual interpretive consistency.

**Relationship to Existing Frameworks.** Our framework complements rather than replaces existing evaluation paradigms. The Holistic Evaluation of Language Models (HELM) framework (Liang et al., 2023) provides infrastructure for multi-metric evaluation including fairness and robustness; we extend this by adding SSH-specific interpretive dimensions. XNLI (Conneau et al., 2018) and related benchmarks provide cross-lingual NLU baselines; our framework adds culturally grounded interpretive tasks. The CAMeL benchmark (Naous et al., 2024) and the cultural alignment evaluations by Sukiennik et al. (2025) provide complementary tools for assessing entity-level cultural bias and value-level cultural alignment respectively; our framework integrates both granularity levels within a unified SSH evaluation pipeline.

**Transparency and Reproducibility.** Researchers must log prompts, outputs, timestamps, model versions, and decoding parameters. For proprietary models, researchers should test for consistency across sessions and document any detected model updates (Linegar et al., 2023). Following Abdurahman et al. (2025), researchers should report disaggregated results across languages and demographic categories represented in texts. Carlson and Burbano (2025) recommend systematic sensitivity analysis across prompt formulations; we adopt this as a core principle: SSH researchers should routinely test at least three prompt variants per annotation task and report the range of results obtained.

## 8. Illustrative Application: Multilingual Political Discourse Analysis

To demonstrate how the proposed framework can be applied in practice, we outline a concrete research scenario involving multilingual political framing analysis, a common task in computational social science.

**Research question.** How do news media in Arabic, English, and French frame immigration policy debates, and can LLMs reliably identify these frames across languages?

**Step 1: Corpus construction.** The researcher assembles native-language editorial articles from major outlets in each language (e.g., Al Jazeera Arabic, The Guardian, Le Monde). Crucially, these are not translations but independently authored texts addressing immigration in their respective national contexts. A Data Statement (Bender and Friedman, 2018) documents language variety (e.g., Modern Standard Arabic vs. Gulf dialect), publication period, editorial stance distribution, and any known ideological affiliations.

**Step 2: Annotation schema development.** Frame categories are developed collaboratively with area specialists in each language community, following Entman's four framing functions (problem definition, causal attribution, moral evaluation, treatment recommendation). The schema is piloted with native-speaker annotators. Suppose inter-annotator reliability yields Krippendorff's $\alpha = 0.74$ for Arabic, $\alpha = 0.82$ for English, and $\alpha = 0.78$ for French. Following Krippendorff (2004), the Arabic and French scores fall in the tentative range (0.67–0.80) and are acceptable for exploratory analysis but would require refinement for confirmatory claims. The English score exceeds 0.80 and supports stronger conclusions.

**Step 3: Cultural alignment check.** Before deploying LLMs for full-corpus annotation, the researcher tests cultural alignment on a probe set of 50 items per language containing culturally specific entities and concepts. Using the Cultural Bias Score method from CAMeL (Naous et al., 2024), the researcher discovers that the model assigns Western-normative immigration framings (e.g., "economic burden") at higher rates in Arabic than Arabic-speaking annotators do, revealing a systematic cultural skew that would contaminate cross-national comparisons if left uncorrected.

**Step 4: Cross-lingual stability assessment.** The researcher identifies a subset of 30 parallel items (events covered in all three languages) and computes pairwise Cohen's $\kappa$ between LLM annotations across languages. If English-French $\kappa = 0.72$ but English-Arabic $\kappa = 0.51$, this asymmetry signals that the model's Arabic frame identification is substantially less stable, likely reflecting weaker Arabic-language priors.

**Step 5: Reasoning faithfulness test.** For a subsample of items, the researcher requests chain-of-thought explanations and applies bias injection (prepending a suggested frame label). If the model's frame assignments shift in 25% of Arabic cases but only 8% of English cases under bias injection, this reveals language-dependent faithfulness, a red flag for cross-lingual interpretive validity.

**Step 6: Reporting.** The researcher documents all findings in the Epistemic Risk Register (Section 9), reports disaggregated metrics, and qualifies cross-lingual conclusions appropriately. The Arabic frame analysis is presented with explicit caveats about reduced stability and cultural alignment, while the English-French comparison receives stronger interpretive weight.

This scenario illustrates that the framework is implementable with modest resources: the probe sets and parallel items are small (50 and 30 items respectively), the statistical tests are standard, and the primary investment is in recruiting area specialists for schema development and cultural probing rather than in large-scale annotation.

## 9. Epistemic and Ethical Implications

The integration of multilingual reasoning LLMs into SSH research introduces epistemic and ethical considerations that extend beyond technical evaluation.

Cultural homogenization risk arises when dominant language priors shape interpretation across contexts. Models trained primarily on English data may impose Western conceptual frameworks on texts from other cultural traditions. Research on cultural alignment reveals that even when generating content in non-English languages, LLMs often reflect the value systems of English-speaking countries (Tao et al., 2024). The empirical findings from Naous et al. (2024) and Sukiennik et al. (2025) demonstrate that this risk is not hypothetical but measurable.

Authority displacement may occur if model outputs are treated as objective rather than mediated. Hayes (2025) cautions that while LLMs offer powerful capabilities for engaging with qualitative data, researchers must maintain critical awareness that model outputs reflect training distributions rather than unmediated access to textual meaning. The risk is amplified when LLMs are used both to generate and evaluate interpretive claims, creating closed loops that may reinforce rather than interrogate model priors (Zheng et al., 2023). Experience from large-scale annotation projects involving culturally sensitive content, such as hate speech annotation, has shown that even human annotators are affected by emotional toll and interpretive fatigue (Al Emadi and Zaghouani, 2024), underscoring the need for careful oversight whether annotation is performed by humans or machines.

Reproducibility challenges emerge when proprietary models are updated without transparency (Bommasani et al., 2021). Open-source models offer greater stability and control, but even these are subject to community-driven updates and version proliferation.

The reliability of survey-based methods for assessing cultural alignment itself warrants critical examination. Recent work has challenged assumptions that cultural alignment is a stable property of models rather than an artifact of evaluation design, and that alignment on one set of cultural dimensions predicts alignment on others. Empirical tests reveal significant instability across presentation formats and incoherence between evaluated and held-out cultural dimensions, reinforcing the need for multi-method evaluation approaches.

We propose an **Epistemic Risk Register** that SSH researchers should complete when deploying multilingual LLMs:

1. *Codebook provenance*: Who defined the annotation categories? Were native speakers from target cultures involved?
2. *Cultural stakeholder engagement*: How were affected communities consulted in evaluation design?
3. *Disaggregation requirements*: What subgroup analyses are required to detect differential performance?
4. *Model update monitoring*: What procedures exist to detect and document model changes over time?
5. *Interpretive authority*: How are model outputs positioned relative to human expert judgment?
6. *Prompt sensitivity documentation*: Were multiple prompt formulations tested, and what was the range of variation in results?
7. *Cultural alignment verification*: Were model outputs assessed for systematic cultural bias using established benchmarks or domain-appropriate cultural probes?

## 10. Toward SSH-Specific Benchmarks

The framework proposed here highlights a significant gap in the current evaluation landscape: the absence of benchmarks specifically designed for SSH interpretive tasks. Existing multilingual benchmarks primarily assess factual knowledge retrieval, natural language inference, or commonsense reasoning. While these capabilities are necessary for SSH applications, they are insufficient.

We identify four properties that SSH-specific benchmarks should exhibit. First, *interpretive pluralism*: tasks should admit multiple defensible answers rather than a single correct response. Evaluation metrics should reward appropriate uncertainty and penalize unwarranted confidence. Second, *cultural grounding*: benchmark items should be developed in their target languages by domain experts embedded in the relevant cultural contexts, not translated from English-language originals. Third, *theoretical anchoring*: annotation categories should derive from established SSH theoretical frameworks rather than ad hoc classification schemes. Fourth, *ecological validity*: test materials should be drawn from the actual text genres that SSH researchers analyze (parliamentary debates, news editorials, social media discourse, interview transcripts) rather than synthetic inputs.

The development of such benchmarks requires sustained collaboration between NLP researchers and SSH scholars. Kuhn (2019) identifies the scheduling dilemma that arises when computational methods require early specification while hermeneutic approaches prefer late specification as understanding develops. We propose that benchmark development adopt an iterative co-design methodology. Community-driven approaches exemplified by Masakhane (Orife et al., 2020) and the Aya initiative (Singh et al., 2024; Üstün et al., 2024) should inform how native speakers and SSH scholars from underrepresented language communities participate in benchmark development as researchers rather than merely as annotators.

## 11. Practical Guidance for Resource-Constrained Settings

A realistic concern, raised in the evaluation of this framework, is that the full protocol may be infeasible for researchers lacking access to large multilingual annotator pools, validated cultural survey data, or extensive computational resources. We address this by proposing a tiered implementation approach.

**Minimum viable evaluation (Tier 1).** Even with limited resources, researchers can implement three basic checks: (a) test at least three prompt variants per task and report variance across formulations, following Carlson and Burbano (2025); (b) compute cross-lingual agreement on a small parallel set (as few as 20 items) to flag gross instability; and (c) inspect a sample of reasoning traces for obvious cultural misalignment using researcher domain expertise. These steps require no specialized infrastructure and can be completed in hours.

**Standard evaluation (Tier 2).** With moderate resources, researchers add: (a) formal inter-annotator reliability with native speakers (minimum two annotators per language, 100 item subsample); (b) cultural bias probing using entity substitution (comparing model behavior on culturally specific vs. Western entities); and (c) bias injection tests on a subsample of reasoning tasks. This tier requires collaborators in each target language community but no large-scale data collection.

**Full protocol (Tier 3).** The complete framework as described in Section 6, including population-level cultural alignment metrics using WVS or ESS data, formal variance decomposition, and comprehensive faithfulness testing. This tier is appropriate for high-stakes research projects with dedicated evaluation budgets.

This tiered approach ensures that even researchers with minimal resources can incorporate epistemic accountability into their workflows, while providing a clear path for scaling evaluation rigor as resources allow.

## 12. Implications for the SSH Research Agenda

Realizing the potential of multilingual reasoning LLMs for SSH research requires developing shared infrastructure for documenting model configurations, logging experimental procedures, and archiving annotation materials. We recommend that SSH researchers adopt Data Statements (Bender and Friedman, 2018) and Model Cards (Mitchell et al., 2019) as minimum documentation standards, extended with the epistemic risk register proposed above.

The field would benefit from establishing multilingual LLM evaluation consortia that pool expertise across language communities and SSH disciplines. Such consortia could maintain living benchmarks that evolve alongside model capabilities, conduct regular cross-model comparative evaluations on SSH-relevant tasks, and develop shared annotation resources that reduce the per-project cost of multilingual validation.

We further recommend that SSH journals and conferences develop reporting standards for studies that use LLM-generated annotations or LLM-assisted analysis. At a minimum, researchers should report the specific model name, version, and access date; the complete prompt text used for each task; the decoding parameters (temperature, top-p); the number of prompt variants tested and the sensitivity of results to prompt choice; disaggregated performance metrics across languages and cultural subgroups; and any detected changes in model behavior over the course of data collection.

## 13. Conclusion

Multilingual reasoning LLMs function as hermeneutic instruments that mediate interpretation in computational social science. Recognizing this role requires moving beyond benchmark accuracy toward epistemically grounded evaluation frameworks that attend to cultural situatedness, reasoning faithfulness, and the structural inequalities embedded in training data. By combining hermeneutics, philosophy of technology, and technical methodology with operationalized metrics, this paper provides a foundation for responsible multilingual reasoning assessment aligned with SSH values. The empirical evidence reviewed here, from cultural bias in Arabic-language models to the unreliability of chain-of-thought reasoning and the instability of cultural alignment evaluations, demonstrates that these concerns are not merely theoretical but have concrete, measurable consequences for the validity of SSH research. The illustrative application in Section 8 demonstrates that meaningful evaluation is achievable even with modest resources when guided by a principled framework. Future work should operationalize these principles in large-scale empirical studies, and we particularly encourage pilot studies that implement the tiered evaluation protocol on small multilingual corpora to demonstrate feasibility and refine metric thresholds. The research community is well-positioned to advance this agenda through sustained interdisciplinary collaboration.

## 14. Limitations

This position paper presents a conceptual and methodological framework for evaluating multilingual reasoning large language models in Social Sciences and Humanities research. The paper is primarily theoretical. The suggested measures for cultural alignment, cross-lingual stability, and reasoning faithfulness have not been validated through large-scale empirical testing across many different models. This is especially true for models that work with low-resource or typologically diverse languages.

The cultural alignment metrics depend on population-level cultural datasets from validated surveys such as the World Values Survey. These datasets may be unavailable, outdated, or insufficiently representative for many non-Western and indigenous groups. The proposed KL divergence threshold of 0.1 nats and the ICC threshold of 0.10 are offered as starting points informed by standard statistical practice, but they require empirical calibration through pilot studies in specific SSH domains before they can be treated as firm benchmarks.

The framework focuses on text-based tasks and does not yet address multimodal analysis or real-time interactive settings. Language-specific challenges such as diverse writing systems, dialectal variation, code-switching, and uneven tokenization may produce larger performance disparities than the current stability metrics capture. The illustrative scenario in Section 8, while designed to be realistic, has not been executed as a full empirical study. Finally, the framework assumes some degree of model access or adaptability, which may not be available with closed commercial models. These points underscore the need for iterative, community-based validation before wide adoption.

## 15. Ethical Considerations

Integrating multilingual reasoning LLMs into SSH research raises ethical concerns beyond epistemic issues. A primary risk is cultural homogenization from English-dominant training data, which perpetuates Western normative interpretations and marginalizes non-Western epistemologies, indigenous knowledge, and minority voices. This reinforces colonial legacies in global knowledge production. Other concerns include deskilling researchers through over-reliance on automation, privacy risks with sensitive qualitative data, misuse of outputs in high-stakes interpretive work without human oversight, and accountability problems from hallucinations or inconsistent cross-lingual performance.

To address these we advocate participatory design with native speakers, Global South scholars, and community stakeholders as co-creators of benchmarks, protocols, and transparency standards, following models like Masakhane (Orife et al., 2020) and Aya (Singh et al., 2024). Researchers must document model provenance, prompts, detected biases, and limitations while keeping human reflexivity central. Responsible adoption requires ongoing ethical reflection to ensure LLMs support rather than replace culturally situated human understanding.

## Acknowledgments

This work was made possible by the National Priorities Research Program grant NPRP14C-0916-210015 from the Qatar National Research Fund (QNRF), part of the Qatar Research, Development and Innovation Council (QRDI). The author also acknowledge the Artificial Intelligence and Media Lab (AIM Lab) at Northwestern University in Qatar (NU-Q) and the MARSAD Lab for providing valuable resources and support that contributed to this research.

## 16. Bibliographical References